\documentclass[letterpaper]{article}
\usepackage[submission]{aaai2026}
\usepackage{times}
\usepackage{helvet}
\usepackage{courier}
\usepackage[hyphens]{url}
\usepackage{graphicx}
\usepackage{natbib}
\usepackage{caption}
\usepackage{amsmath,amssymb}
\usepackage{booktabs}
\usepackage{algorithm}
\usepackage{algorithmic}
\usepackage{xcolor}
\usepackage{multirow}

\title{Generic Expert Coverage for Pruning Sparse\\Mixture-of-Experts Language Models}

\author{Yongqin Zeng, Sicheng Pan, Jiale Wang, Hai-tao Zheng, Hong-Gee Kim, Chunxia Ma, XiuTeng Zhou}
\affiliations{}

\begin{document}
\maketitle

\begin{abstract}
Sparsely activated Mixture-of-Experts (MoE) language models contain substantial structured redundancy among routed experts, but pruning them without downstream calibration data remains challenging.
Existing expert-pruning methods typically rely on a single aggregated importance score, which can bias the retained set toward experts favored by dominant calibration patterns.
We propose \textbf{Generic TB-Coverage}, a coverage-aware expert pruning method that uses only generic text corpora (WikiText2 and C4) for calibration.
Instead of collapsing expert utility into one score, our method profiles per-expert utility separately on each corpus and enforces a fixed-budget coverage rule that preserves high-utility experts from each corpus before constructing the final pruning mask.
Across Qwen1.5-MoE-A2.7B and DeepSeek-MoE-16B-Base at 25\%, 50\%, and 75\% retention budgets, our method improves average accuracy on six common zero-shot benchmarks over random pruning, REAP, and ExpertSparsity, while also reducing perplexity degradation on WikiText2 and C4.
The gains are largest under aggressive pruning (25\% and 50\% retain), suggesting that preserving cross-corpus expert coverage is an effective generic-data prior for MoE pruning.
Our improvements hold with fixed pruning budgets and no downstream calibration data.
\end{abstract}

\section{Introduction}

Sparse Mixture-of-Experts (MoE) language models improve parameter efficiency by activating only a small subset of experts per token, achieving strong performance while keeping per-token computation manageable~\citep{shazeer2017outrageously,fedus2022switch,jiang2024mixtral}.
This sparse activation also creates structured redundancy: many experts can be removed with limited quality loss if the retained subset preserves the model's core behaviors~\citep{dai2024deepseekmoe}.
The challenge is selecting that subset without access to downstream validation data.

Existing expert-pruning methods usually rank experts using a single scalar criterion, such as routing frequency, reconstruction utility, or REAP-style importance~\citep{lu2024expertsparsity}.
However, when calibration data are heterogeneous, scalar aggregation can over-favor experts that dominate the average signal and under-retain experts that support less frequent but still important generic language behaviors.
For example, consider two generic calibration corpora: WikiText2 (encyclopedic, narrative text) and C4 (broad web text).
An expert that contributes strongly on encyclopedic patterns but not on web discourse will receive a moderate average score, even though it may be the \emph{top-ranked} expert on WikiText2.
A scalar ranking would likely discard it in favor of experts that score moderately on both corpora but are not critical for either.
This suggests that expert retention should preserve cross-corpus coverage rather than only optimize a single aggregated importance score.

This problem is especially acute when the goal is \emph{general-purpose pruning} without downstream calibration data.
A principled pruning method should not rely on downstream evaluation tasks during calibration, since good downstream numbers could then partly reflect task leakage rather than genuine generalization.
We study expert pruning in this practically important setting: \emph{can we prune MoE experts using only generic language corpora while preserving broad downstream behavior?}

Our key observation is that even purely generic corpora do not exercise the same experts in the same way.
A pruning rule that preserves only globally dominant experts can therefore reduce expert coverage across generic behaviors before downstream transfer is ever evaluated.
We address this by profiling expert utility separately on multiple generic corpora and enforcing balanced protection across corpus-specific rankings.

We propose \textbf{Generic TB-Coverage} (Task-Balanced Coverage), a coverage-aware expert pruning method.
The method profiles per-expert importance separately for each generic calibration corpus, builds corpus-specific expert rankings, and selects protected experts through a round-robin coverage rule.
Protected experts are then merged into a reconstruction-stable candidate mask, and the exact retention budget is restored by removing the lowest-ranked unprotected experts.
The entire procedure uses only WikiText2 and C4 for calibration and requires no fine-tuning.

We evaluate on two MoE language models---Qwen1.5-MoE-A2.7B~\citep{bai2024qwen} and DeepSeek-MoE-16B-Base~\citep{dai2024deepseekmoe}---at three expert-retention budgets (25\%, 50\%, 75\%).
The primary metric is the average accuracy over six common zero-shot benchmarks (ARC-Challenge, ARC-Easy, HellaSwag, PIQA, WinoGrande, BoolQ); we additionally report MMLU, GSM8K, and Math500 as auxiliary stress tests.

Our contributions are:
\begin{enumerate}
\item We identify a failure mode of scalar expert ranking for downstream-free MoE pruning: dominant calibration patterns can over-concentrate the retained expert set.
\item We propose a simple coverage-aware pruning rule that preserves high-utility experts across multiple generic corpora before fixed-budget mask construction.
\item We show that this rule improves pruning quality on two open MoE LMs across three retention budgets and six zero-shot benchmarks.
\item We analyze random-pruning variance and show that single-seed random baselines can be misleading in sparse MoE pruning.
\end{enumerate}

\section{Related Work}

\textbf{MoE model compression.}
Sparse MoE architectures~\citep{shazeer2017outrageously,lepikhin2021gshard,fedus2022switch} route each token to a small subset of experts, creating natural opportunities for expert-level pruning.
ST-MoE~\citep{zoph2022stmoe} studies training stability in sparse models, while DeepSeek-MoE~\citep{dai2024deepseekmoe} introduces fine-grained expert segmentation that increases expert count and hence pruning potential.
MoEfication~\citep{fu2023moefication} converts dense feed-forward layers into MoE structures.

\textbf{Expert importance criteria.}
Several criteria have been proposed for ranking expert importance.
Routing frequency counts how often each expert is selected by the router.
REAP-style methods compute importance scores that combine routing probability with expert output magnitude.
ExpertSparsity~\citep{lu2024expertsparsity} uses reconstruction-based layerwise pruning, selecting experts to minimize the reconstruction error between the pruned and original MoE layer outputs.
These methods share the limitation that they optimize a single scalar objective, which can cause over-concentration of retained experts.

\textbf{LLM pruning.}
Beyond expert pruning, structured pruning for large language models includes layer removal~\citep{men2024shortgpt}, width pruning~\citep{ma2023llmpruner}, and unstructured sparsity~\citep{frantar2023sparsegpt,sun2024simple}.
SliceGPT~\citep{ashkboos2024slicegpt} removes components via singular value decomposition.
These methods operate at the weight or layer level rather than the expert level and are complementary to our approach.

\textbf{Coverage and diversity in pruning.}
The idea of preserving diversity in pruned models has been explored in network pruning for convolutional networks, where filter diversity criteria help maintain representational capacity.
To our knowledge, Generic TB-Coverage is the first to explicitly introduce a coverage rule for expert-level MoE pruning based on per-corpus profiling.

\section{Method}

\subsection{Problem Formulation}

Consider a sparse MoE language model with $L$ MoE layers.
At each MoE layer $l$, the model has $N$ routed experts $\{e_1, \ldots, e_N\}$.
For each input token with hidden state $h_x$, the router selects the top-$k$ experts and computes a weighted sum of their outputs:
\begin{equation}
\text{MoE}(h_x) = \sum_{e \in \text{TopK}(h_x)} g_e(x) \cdot f_e(h_x),
\end{equation}
where $g_e(x)$ is the router's softmax probability assigned to expert $e$ for token $x$ (set to zero for non-selected experts), and $f_e(h_x)$ is the expert's feed-forward output on the hidden state $h_x$.

Given a retain ratio $\rho \in (0, 1]$, the pruning budget retains $K = \lfloor \rho N \rfloor$ routed experts per MoE layer.
The goal is to construct a binary expert mask $m_l \in \{0,1\}^N$ with exactly $K$ ones per layer such that the pruned model preserves broad language performance.
After pruning, the router selects among the retained experts only.

\subsection{Generic Expert Profiling}

The method uses two generic calibration corpora: WikiText2~\citep{merity2016wikitext} and C4~\citep{raffel2020c4}.
For each corpus $t \in \mathcal{T} = \{\text{WikiText2}, \text{C4}\}$, we run the model on calibration text and collect per-expert statistics independently for each MoE layer.

For each layer $l$, expert $e$, and calibration corpus $t$, we compute a REAP-style utility score that combines the router weight with the expert output norm, averaged over all calibration tokens where expert $e$ is selected by the router:
\begin{equation}
\label{eq:score}
s_t^{(l)}(e) = \mathbb{E}_{x \sim \mathcal{D}_t}\left[ g_e(x) \cdot \| f_e(h_x^{(l)}) \|_2 \;\big|\; e \in \text{TopK}(x) \right],
\end{equation}
where $h_x^{(l)}$ is the hidden state at layer $l$, $g_e(x)$ is the router weight for expert $e$ on token $x$, and $f_e(h_x^{(l)})$ is the expert output.
The expectation is approximated by averaging over all tokens in the calibration set from corpus $t$ for which expert $e$ is among the top-$k$ selected experts.
This score captures both how strongly the router activates expert $e$ (through $g_e$) and how much it contributes to the hidden state (through $\|f_e\|_2$).
We profile each calibration corpus independently, producing a separate score vector $\mathbf{s}_t^{(l)} = [s_t^{(l)}(e_1), \ldots, s_t^{(l)}(e_N)]$ per MoE layer.

\subsection{Coverage-Aware Expert Protection}

Instead of ranking experts by their average score $\bar{s}^{(l)}(e) = \frac{1}{|\mathcal{T}|}\sum_{t} s_t^{(l)}(e)$, Generic TB-Coverage builds per-corpus rankings and selects protected experts by round-robin across corpora.

For each MoE layer $l$, we define a per-layer protection budget $B$ (a hyperparameter controlling how many experts receive coverage protection), chosen such that $B \leq K$ for every layer and retention budget, guaranteeing feasibility.
The round-robin procedure works as follows:
\begin{enumerate}
\item Sort all $N$ experts by $s_{\text{WikiText2}}^{(l)}(e)$ in descending order, producing ranking $\pi_{\text{Wiki}}$.
\item Sort all $N$ experts by $s_{\text{C4}}^{(l)}(e)$ in descending order, producing ranking $\pi_{\text{C4}}$.
\item Starting with an empty protected set $P_l$, alternate between the two rankings: at each step, take the highest-ranked expert from the current corpus's ranking that is not yet in $P_l$, and add it.
\item Stop when $|P_l| = B$.
\end{enumerate}

This ensures that both corpora contribute to the protected set regardless of the absolute score magnitudes.
An expert that is top-ranked on WikiText2 but mid-ranked on C4 will still be protected, even though its average score might not place it in the top-$K$ by a single-criterion ranking.
If both rankings agree on the top experts (high overlap), the round-robin degenerates to standard top-$B$ selection, so the rule introduces no harm in that case.

\subsection{Budget-Preserving Mask Construction}

\begin{algorithm}[t]
\caption{Generic TB-Coverage Expert Pruning}
\label{alg:method}
\textbf{Input}: MoE model $\mathcal{M}$; corpora $\mathcal{T} = \{\text{WikiText2}, \text{C4}\}$; retain ratio $\rho$; protection budget $B$\\
\textbf{Output}: Mask $m$ with $K = \lfloor \rho N \rfloor$ retained experts per layer
\begin{algorithmic}[1]
\FOR{each MoE layer $l$}
  \FOR{each corpus $t \in \mathcal{T}$}
    \STATE Profile $\mathcal{M}$ on $t$; compute scores $s_t(e)$
  \ENDFOR
  \STATE Build per-corpus rankings by $s_t(e)$
  \STATE Select $P_l$ by round-robin over rankings ($|P_l| = B$)
  \STATE Init mask from reconstruction candidate $\hat{m}_l$
  \STATE Merge: set $m_l(e) = 1$ for all $e \in P_l$
  \IF{$\sum_e m_l(e) > K$}
    \STATE Remove lowest-ranked unprotected by $\bar{s}^{(l)}$ until $K$ met
  \ENDIF
\ENDFOR
\RETURN mask $m$
\end{algorithmic}
\end{algorithm}

The final mask keeps exactly $K$ experts per MoE layer.
We initialize from a reconstruction-stable candidate mask $\hat{m}_l$ (obtained via layerwise reconstruction minimization on C4) that already retains $K$ experts per layer.
We merge the protected set $P_l$ into this candidate: set $m_l(e) = 1$ for all $e \in P_l$ (protected experts are always retained), then if the union exceeds $K$, remove the lowest-ranked \emph{unprotected} experts in order of ascending average score $\bar{s}^{(l)}(e)$ until exactly $K$ experts remain.
Protected experts are never removed; the budget is restored entirely by dropping unprotected reconstruction-selected experts.

This procedure is budget-preserving: the final mask always retains exactly $K$ experts per MoE layer.
Algorithm~\ref{alg:method} summarizes the full method.

\subsection{Complexity Analysis}

With $|\mathcal{T}|$ calibration corpora, the additional profiling cost is $|\mathcal{T}|$ forward-only passes over the calibration set (no gradient computation).
With two corpora and approximately 512 sequences of length 1024 tokens, this is modest relative to model training.
Mask construction involves per-layer sorting of $N$ experts ($O(N \log N)$ per layer) and round-robin selection ($O(B)$ per layer), which is negligible relative to the profiling cost.
The method stores only per-expert scalar scores per layer and does not require post-pruning fine-tuning.

\section{Experiments}

\subsection{Setup}

\textbf{Models.}
We evaluate on two sparse MoE language models:
\begin{itemize}
\item \textbf{Qwen1.5-MoE-A2.7B}~\citep{bai2024qwen}: 60 routed experts per MoE layer, with top-4 routing.
\item \textbf{DeepSeek-MoE-16B-Base}~\citep{dai2024deepseekmoe}: 64 routed experts per MoE layer, with top-6 routing.
\end{itemize}

\textbf{Retain budgets.}
We report three expert-retention ratios: 25\%, 50\%, and 75\%.
For Qwen, this corresponds to retaining 15, 30, or 45 out of 60 routed experts per layer.
For DeepSeek, this corresponds to retaining 16, 32, or 48 out of 64 routed experts per layer.

\textbf{Calibration.}
Our method uses WikiText2~\citep{merity2016wikitext} and C4~\citep{raffel2020c4} only.
We draw 512 sequences of length 1024 tokens from each corpus.
No downstream benchmark data is used during calibration.

\textbf{Baselines.}
We compare against three baselines:
\begin{itemize}
\item \textbf{Random pruning}: Uniform random expert removal per layer, reported as mean and standard deviation over six seeds (0, 1, 2, 3, 4, 42).
\item \textbf{Original REAP}: Direct top-$k$ expert selection using REAP-style importance scores. For a fair comparison with a method that does use task-specific calibration, we run REAP with Evol-CodeAlpaca calibration (128 texts, sequence length 2048).
\item \textbf{ExpertSparsity}~\citep{lu2024expertsparsity}: Reconstruction-based layerwise expert pruning using C4, adapted to our model wrappers. We follow the original paper's layerwise reconstruction protocol.
\end{itemize}
We note that REAP uses domain-specific calibration data while our method uses only generic corpora, making the comparison favorable to REAP in terms of calibration informativeness.

\textbf{Evaluation.}
We evaluate on two language modeling metrics (WikiText2 PPL, C4 PPL, computed on the standard validation splits using sliding-window evaluation) and six primary downstream tasks: ARC-Challenge and ARC-Easy~\citep{clark2018arc}, HellaSwag~\citep{zellers2019hellaswag}, PIQA~\citep{bisk2020piqa}, WinoGrande~\citep{sakaguchi2021winogrande}, and BoolQ~\citep{clark2019boolq}.
We report \emph{Common Avg} as the unweighted average over these six tasks.
We additionally report MMLU~\citep{hendrycks2021mmlu}, GSM8K~\citep{cobbe2021gsm8k}, and Math500~\citep{lightman2023math} as auxiliary stress tests for knowledge and reasoning.
All evaluations use full validation/test sets with zero-shot prompting (no few-shot examples).

\subsection{Main Results}

Table~\ref{tab:main} presents the main results.
Generic TB-Coverage achieves the highest Common Avg across both models and all three retain budgets.

\begin{table*}[t]
\centering
\caption{Main results on Qwen1.5-MoE-A2.7B and DeepSeek-MoE-16B-Base. Random pruning reports mean $\pm$ std over six seeds. Other methods are deterministic. \textbf{Bold}: best Common Avg per group.}
\label{tab:main}
\small
\setlength{\tabcolsep}{2pt}
\begin{tabular}{lllcccccc}
\toprule
Model & Ret. & Method & Avg & MMLU & GSM & M500 & W.PPL & C.PPL \\
\midrule
\multirow{4}{*}{Qwen} & \multirow{4}{*}{25\%}
  & Random & 0.417$\pm$0.021 & 0.240$\pm$0.011 & --- & --- & 920.67$\pm$744.23 & 3109.75$\pm$2197.97 \\
  & & Original REAP & 0.438 & 0.247 & 0.000 & 0.000 & 193.65 & 551.91 \\
  & & Paper ExpertSparsity & 0.407 & 0.233 & 0.000 & 0.000 & 160.93 & 358.90 \\
  & & \textbf{Generic TB-Coverage} & \textbf{0.448} & 0.234 & 0.000 & 0.000 & \textbf{146.22} & \textbf{259.12} \\
\midrule
\multirow{4}{*}{Qwen} & \multirow{4}{*}{50\%}
  & Random & 0.535$\pm$0.060 & 0.315$\pm$0.074 & 0.000$\pm$0.000 & 0.002$\pm$0.002 & 77.95$\pm$93.70 & 165.33$\pm$178.95 \\
  & & Original REAP & 0.552 & 0.406 & 0.000 & 0.002 & 23.77 & 61.47 \\
  & & Paper ExpertSparsity & 0.519 & 0.325 & 0.000 & 0.000 & 36.18 & 88.31 \\
  & & \textbf{Generic TB-Coverage} & \textbf{0.598} & 0.369 & 0.000 & 0.004 & \textbf{16.76} & \textbf{47.01} \\
\midrule
\multirow{4}{*}{Qwen} & \multirow{4}{*}{75\%}
  & Random & 0.645$\pm$0.041 & 0.516$\pm$0.035 & 0.000$\pm$0.000 & 0.003$\pm$0.004 & 13.25$\pm$1.49 & 33.80$\pm$5.17 \\
  & & Original REAP & 0.629 & 0.519 & 0.000 & 0.002 & 13.78 & 32.81 \\
  & & Paper ExpertSparsity & 0.655 & 0.548 & 0.000 & 0.002 & 11.92 & 29.41 \\
  & & \textbf{Generic TB-Coverage} & \textbf{0.662} & 0.528 & 0.002 & 0.002 & \textbf{10.57} & \textbf{28.22} \\
\midrule
\multirow{4}{*}{DeepSeek} & \multirow{4}{*}{25\%}
  & Random & 0.403$\pm$0.025 & 0.243$\pm$0.010 & --- & --- & 728.88$\pm$818.42 & 2640.77$\pm$2971.07 \\
  & & Original REAP & 0.395 & 0.238 & 0.000 & 0.006 & 544.15 & 1734.92 \\
  & & Paper ExpertSparsity & 0.404 & 0.240 & 0.000 & 0.000 & 176.63 & 615.18 \\
  & & \textbf{Generic TB-Coverage} & \textbf{0.448} & 0.232 & 0.000 & 0.000 & \textbf{137.03} & \textbf{423.72} \\
\midrule
\multirow{4}{*}{DeepSeek} & \multirow{4}{*}{50\%}
  & Random & 0.488$\pm$0.082 & 0.244$\pm$0.010 & 0.000$\pm$0.000 & 0.005$\pm$0.006 & 54.19$\pm$51.78 & 249.73$\pm$311.32 \\
  & & Original REAP & 0.542 & 0.287 & 0.000 & 0.008 & 27.33 & 61.60 \\
  & & Paper ExpertSparsity & 0.571 & 0.264 & 0.000 & 0.010 & 20.88 & 53.47 \\
  & & \textbf{Generic TB-Coverage} & \textbf{0.606} & 0.242 & 0.000 & 0.006 & \textbf{16.43} & \textbf{45.14} \\
\midrule
\multirow{4}{*}{DeepSeek} & \multirow{4}{*}{75\%}
  & Random & 0.568$\pm$0.116 & 0.260$\pm$0.019 & 0.000$\pm$0.000 & 0.006$\pm$0.009 & 19.19$\pm$9.55 & 80.94$\pm$81.39 \\
  & & Original REAP & 0.649 & 0.340 & 0.000 & 0.000 & 14.08 & 29.94 \\
  & & Paper ExpertSparsity & 0.678 & 0.314 & 0.000 & 0.004 & 12.17 & 27.70 \\
  & & \textbf{Generic TB-Coverage} & \textbf{0.683} & 0.309 & 0.000 & 0.006 & \textbf{10.12} & \textbf{25.96} \\
\bottomrule
\end{tabular}
\end{table*}

\begin{figure}[t]
\centering
\includegraphics[width=0.95\columnwidth]{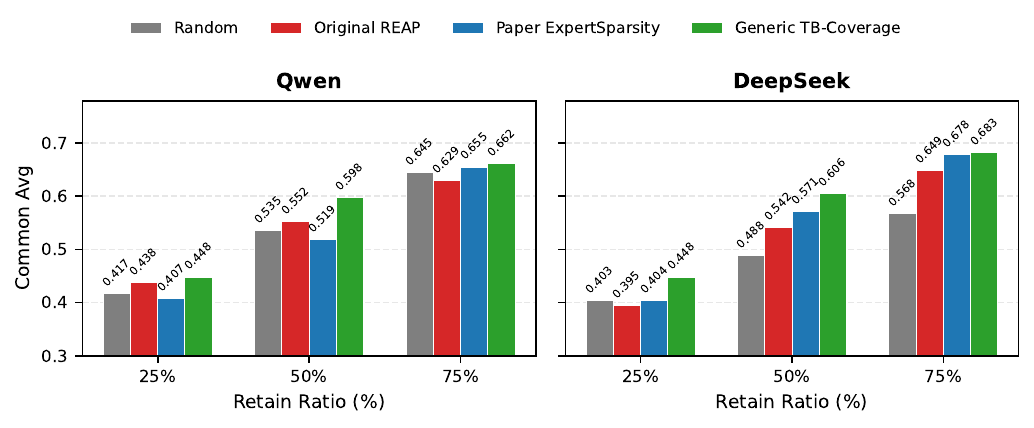}
\caption{Common Avg accuracy across methods and retain ratios on both models. Generic TB-Coverage (green) consistently achieves the highest average accuracy across all settings.}
\label{fig:common_avg}
\end{figure}

On Qwen1.5-MoE-A2.7B, Generic TB-Coverage improves Common Avg over Paper ExpertSparsity by +0.041, +0.080, and +0.007 at 25\%, 50\%, and 75\% retain, respectively.
On DeepSeek-MoE-16B-Base, the improvements are +0.044, +0.035, and +0.006.
The gains are largest under aggressive pruning (25\% and 50\% retain), where retaining the right experts matters most.
At 75\% retain, all methods perform closer together since only a quarter of experts are removed.

\begin{figure}[t]
\centering
\includegraphics[width=0.95\columnwidth]{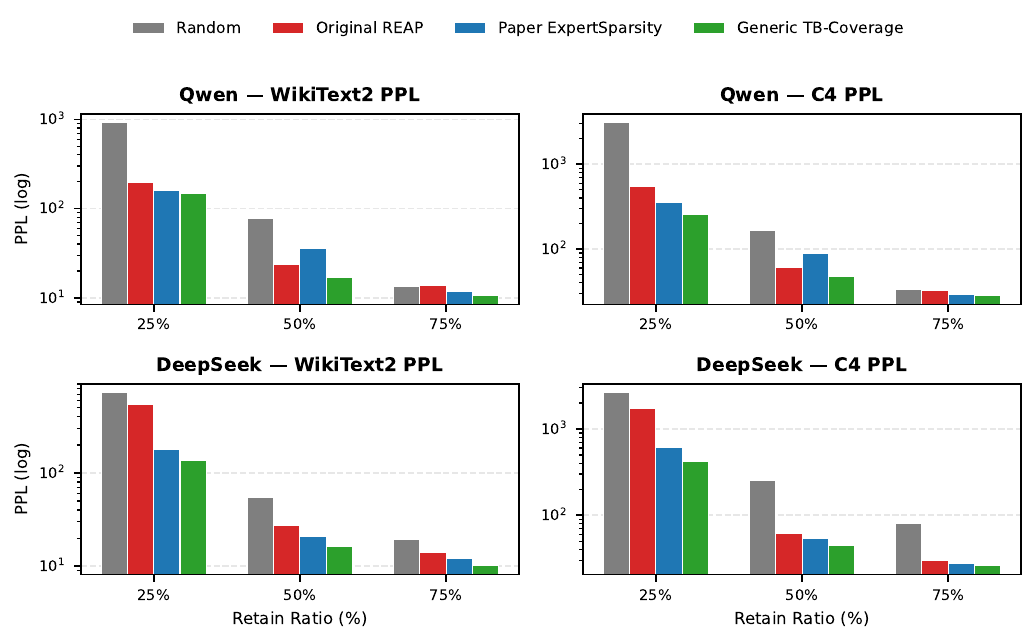}
\caption{WikiText2 and C4 perplexity (log scale, lower is better) across methods and retain ratios. Generic TB-Coverage achieves substantially lower PPL than all baselines, with the largest gains under aggressive pruning.}
\label{fig:ppl}
\end{figure}

Figure~\ref{fig:common_avg} visualizes the Common Avg comparison, and Figure~\ref{fig:ppl} shows the PPL results.
The perplexity improvements are substantial.
For DeepSeek at 25\% retain, C4 PPL drops from 1734.92 (Original REAP) and 615.18 (Paper ExpertSparsity) to 423.72 under Generic TB-Coverage.
Similarly, WikiText2 PPL drops from 544.15 and 176.63 to 137.03.
These reductions indicate that coverage-aware expert selection not only preserves downstream task accuracy but also stabilizes the language modeling distribution.

\subsection{Random Multi-Seed Analysis}

Random pruning is commonly used as a baseline, but single-seed results can be misleading.
Table~\ref{tab:random} shows that random pruning exhibits high variance across seeds.

\begin{table}[t]
\centering
\caption{Random pruning (mean $\pm$ std over 6 seeds) vs.\ Generic TB-Coverage. Single-seed random results can appear competitive but are not reliable.}
\label{tab:random}
\small
\setlength{\tabcolsep}{2pt}
\begin{tabular}{lllcc}
\toprule
Model & Retain & Metric & Random & Ours \\
\midrule
\multirow{3}{*}{Qwen} & \multirow{3}{*}{25\%}
  & Common Avg & 0.417$\pm$0.021 & 0.448 \\
  & & Wiki PPL & 920.67$\pm$744.23 & 146.22 \\
  & & C4 PPL & 3109.75$\pm$2197.97 & 259.12 \\
\midrule
\multirow{3}{*}{Qwen} & \multirow{3}{*}{50\%}
  & Common Avg & 0.535$\pm$0.060 & 0.598 \\
  & & Wiki PPL & 77.95$\pm$93.70 & 16.76 \\
  & & C4 PPL & 165.33$\pm$178.95 & 47.01 \\
\midrule
\multirow{3}{*}{Qwen} & \multirow{3}{*}{75\%}
  & Common Avg & 0.645$\pm$0.041 & 0.662 \\
  & & Wiki PPL & 13.25$\pm$1.49 & 10.57 \\
  & & C4 PPL & 33.80$\pm$5.17 & 28.22 \\
\midrule
\multirow{3}{*}{DeepSeek} & \multirow{3}{*}{25\%}
  & Common Avg & 0.403$\pm$0.025 & 0.448 \\
  & & Wiki PPL & 728.88$\pm$818.42 & 137.03 \\
  & & C4 PPL & 2640.77$\pm$2971.07 & 423.72 \\
\midrule
\multirow{3}{*}{DeepSeek} & \multirow{3}{*}{50\%}
  & Common Avg & 0.488$\pm$0.082 & 0.606 \\
  & & Wiki PPL & 54.19$\pm$51.78 & 16.43 \\
  & & C4 PPL & 249.73$\pm$311.32 & 45.14 \\
\midrule
\multirow{3}{*}{DeepSeek} & \multirow{3}{*}{75\%}
  & Common Avg & 0.568$\pm$0.116 & 0.683 \\
  & & Wiki PPL & 19.19$\pm$9.55 & 10.12 \\
  & & C4 PPL & 80.94$\pm$81.39 & 25.96 \\
\bottomrule
\end{tabular}
\end{table}

At 75\% retain on Qwen, a single random seed (seed 42) achieves Common Avg of 0.669, which appears competitive with our method's 0.662.
However, the multi-seed mean is 0.645$\pm$0.041, and the worst seed drops to 0.604.
Similarly, on DeepSeek at 75\%, random PPL ranges from 10.7 to 31.6 across seeds.
This high variance demonstrates that single-seed random pruning is not a reliable baseline and that principled coverage-aware selection is necessary.

\subsection{Discussion}

\textbf{Gains under aggressive pruning.}
The improvement from Generic TB-Coverage is largest at 25\% and 50\% retain, where retaining the right experts is critical.
At 75\% retain, most experts are preserved and the gap between methods narrows.
This pattern is consistent across both models and suggests that coverage-aware selection is most valuable when the pruning budget is tight.
Our method addresses selection bias in expert retention, but does not optimize routing adaptation after pruning.

\textbf{PPL stability.}
Generic TB-Coverage achieves the lowest perplexity in every setting.
The PPL improvements are especially large at 25\% retain, where the method reduces Wiki PPL by 9\%--24\% and C4 PPL by 28\%--39\% over the next-best baseline (ExpertSparsity).
The absolute PPL values at 25\% retain are high (e.g., C4 PPL of 423.72 for DeepSeek), which reflects the severity of aggressive pruning rather than an evaluation artifact.
Under aggressive pruning, the language model distribution degrades substantially for all methods, and our method degrades least.

\textbf{Reasoning benchmarks.}
GSM8K scores are near zero across all methods and retain budgets, and Math500 scores are similarly low.
We include these results as stress tests but do not claim that Generic TB-Coverage preserves mathematical reasoning ability.
The low scores likely reflect the limited reasoning capacity of the base models rather than a shortcoming of the pruning method.

\textbf{Per-corpus protection budget.}
The current experiments use fixed protection budgets (20 experts for Qwen, 24 for DeepSeek) across all retain ratios.
These values were selected to provide sufficient coverage without dominating the mask at high retain ratios.
Sensitivity analysis of the protection budget is an important direction for future work.
We note that the coverage rule's benefit could be further isolated by comparing against simple multi-corpus aggregation baselines (e.g., mean or max of per-corpus scores); we leave this systematic ablation to future work.

\section{Conclusion}

We have presented Generic TB-Coverage, a coverage-aware expert pruning method for sparse MoE language models.
By profiling expert utility separately on WikiText2 and C4 and protecting experts through a round-robin coverage rule, the method preserves generic language behaviors that single-criterion methods may discard.
Across two MoE models and three retain budgets, Generic TB-Coverage improves downstream average accuracy and language modeling perplexity over random pruning, direct REAP, and reconstruction-based ExpertSparsity, with the largest gains under aggressive pruning.

\textbf{Limitations.}
Our method is intentionally simple and static: it does not adapt experts after pruning, learn corpus weights, or optimize a formal diversity objective.
The current study uses only two generic corpora and two open-base MoE models of moderate scale (2.7B and 16B parameters); whether the same coverage rule scales to instruction-tuned models, larger expert counts, or pruning-plus-quantization settings remains open.
GSM8K and Math500 scores are near zero across all methods, so we cannot assess the method's effect on reasoning ability.
The protection budget $B$ is fixed across retain ratios and requires manual selection.
Finally, we do not measure end-to-end inference latency or memory footprint after pruning.

\bibliography{references}

@article{fedus2022switch,
  title={Switch Transformers: Scaling to Trillion Parameter Models with Simple and Efficient Sparsity},
  author={Fedus, William and Zoph, Barret and Shazeer, Noam},
  journal={Journal of Machine Learning Research},
  volume={23},
  number={120},
  pages={1--39},
  year={2022}
}

@misc{jiang2024mixtral,
  title={Mixtral of Experts},
  author={Jiang, Albert Q and Sablayrolles, Alexandre and Roux, Antoine and Mensch, Arthur and Savary, Blanche and Bamford, Chris and Chaplot, Devendra Singh and Casas, Diego de las and Hanna, Emma Bou and Bressand, Florian and others},
  journal={arXiv preprint arXiv:2401.04088},
  year={2024}
}

@misc{bai2024qwen,
  title={Qwen1.5-MoE: Matching 7B Model Performance with 1/3 Activated Parameters},
  author={Bai, Jinze and Bai, Shuai and Chu, Yunfei and Cui, Zeyu and Dang, Kai and Deng, Xiaodong and Fan, Yang and Ge, Wenbin and Han, Yu and Huang, Fei and others},
  journal={arXiv preprint arXiv:2401.11340},
  year={2024}
}

@inproceedings{dai2024deepseekmoe,
  title={DeepSeek-{MoE}: Towards Ultimate Expert Specialization in Mixture-of-Experts Language Models},
  author={Dai, Damai and Deng, Chengqi and Zhao, Chenggang and Xu, R. X. and Gao, Huazuo and Chen, Deli and Li, Jie and Zeng, Wangding and Zhang, Xing and Wang, Yu and others},
  booktitle={Proceedings of the 41st International Conference on Machine Learning},
  year={2024}
}

@inproceedings{lu2024expertsparsity,
  title={Not All Experts are Equal: Efficient Expert Pruning and Skipping for Mixture of Experts Large Language Models},
  author={Lu, Xudong and Huang, Aojun and Liu, Yuxuan and Qiu, Weikai and Zhou, Lei and Li, Jianlin and Bian, Jiang and Li, Ge and Li, Zhiming},
  booktitle={Proceedings of the 2024 Conference on Empirical Methods in Natural Language Processing},
  year={2024}
}

@inproceedings{ashkboos2024slicegpt,
  title={Slice{GPT}: Compress Large Language Models by Deleting and Optimizing Layers},
  author={Ashkboos, Saleh and Croci, Max and Nascimento, Marcelo Gennari do and Hensman, James and James, Dan and Hoeche, Stefan},
  booktitle={Proceedings of the 41st International Conference on Machine Learning},
  year={2024}
}

@inproceedings{men2024shortgpt,
  title={Short{GPT}: Layers in Large Language Models are More Redundant Than You Expect},
  author={Men, Xin and He, Mingyu and Xu, Qingyu and Wang, Yujin and Luo, Bingbing and Zhang, Min},
  booktitle={Proceedings of the 62nd Annual Meeting of the Association for Computational Linguistics},
  year={2024}
}

@inproceedings{ma2023llmpruner,
  title={LLM-Pruner: On the Structural Pruning of Large Language Models},
  author={Ma, Xinyin and Fang, Gongfan and Wang, Xinchao},
  booktitle={Advances in Neural Information Processing Systems},
  year={2023}
}

@inproceedings{merity2016wikitext,
  title={Pointer Sentinel Mixture Models},
  author={Merity, Stephen and Xiong, Caiming and Bradbury, James and Socher, Richard},
  booktitle={Proceedings of the 5th International Conference on Learning Representations},
  year={2017}
}

@inproceedings{raffel2020c4,
  title={Exploring the Limits of Transfer Learning with a Unified Text-to-Text Transformer},
  author={Raffel, Colin and Shazeer, Noam and Roberts, Adam and Lee, Katherine and Narang, Sharan and Matena, Michael and Zhou, Yanqi and Li, Wei and Liu, Peter J},
  journal={Journal of Machine Learning Research},
  volume={21},
  number={140},
  pages={1--67},
  year={2020}
}

@inproceedings{clark2018arc,
  title={Think You Have Solved Question Answering? Try {ARC}, the {AI2} Reasoning Challenge},
  author={Clark, Peter and Cowhey, Isaac and Etzioni, Oren and Khot, Tushar and Sabharwal, Ashish and Schoenick, Carissa and Tafjord, Oyvind},
  journal={arXiv preprint arXiv:1803.05457},
  year={2018}
}

@inproceedings{zellers2019hellaswag,
  title={{HellaSwag}: Can a Machine Really Finish Your Sentence?},
  author={Zellers, Rowan and Holtzman, Ari and Bisk, Yonatan and Farhadi, Ali and Choi, Yejin},
  booktitle={Proceedings of the 57th Annual Meeting of the Association for Computational Linguistics},
  year={2019}
}

@inproceedings{bisk2020piqa,
  title={{PIQA}: Reasoning about Physical Commonsense in Natural Language},
  author={Bisk, Yonatan and Zellers, Rowan and Gao, Jianfeng and Choi, Yejin},
  booktitle={Proceedings of the AAAI Conference on Artificial Intelligence},
  volume={34},
  number={05},
  pages={7432--7439},
  year={2020}
}

@inproceedings{sakaguchi2021winogrande,
  title={{WinoGrande}: An Adversarial Winograd Schema Challenge at Scale},
  author={Sakaguchi, Keisuke and Le Bras, Ronan and Bhagavatula, Chandra and Choi, Yejin},
  journal={Communications of the ACM},
  volume={64},
  number={9},
  pages={99--106},
  year={2021}
}

@inproceedings{clark2019boolq,
  title={{BoolQ}: Exploring the Surprising Difficulty of Natural Yes/No Questions},
  author={Clark, Christopher and Lee, Kenton and Chang, Ming-Wei and Kwiatkowski, Tom and Collins, Michael and Toutanova, Kristina},
  booktitle={Proceedings of the 2019 Conference of the North American Chapter of the Association for Computational Linguistics},
  year={2019}
}

@inproceedings{hendrycks2021mmlu,
  title={Measuring Massive Multitask Language Understanding},
  author={Hendrycks, Dan and Burns, Collin and Basart, Steven and Zou, Andy and Mazeika, Mantas and Song, Dawn and Steinhardt, Jacob},
  booktitle={Proceedings of the International Conference on Learning Representations},
  year={2021}
}

@inproceedings{cobbe2021gsm8k,
  title={Training Verifiers to Solve Math Word Problems},
  author={Cobbe, Karl and Kosaraju, Vineet and Bavarian, Mohammad and Chen, Mark and Jun, Heewoo and Kaiser, Lukasz and Plappert, Matthias and Tworek, Jerry and Hilton, Jacob and Nakano, Reiichiro and Hesse, Christopher and Schulman, John},
  journal={arXiv preprint arXiv:2110.14168},
  year={2021}
}

@misc{lightman2023math,
  title={Let's Verify Step by Step},
  author={Lightman, Hunter and Kosaraju, Vineet and Burda, Yuri and Edwards, Harri and Baker, Bowen and Lee, Teddy and Leike, Jan and Schulman, John and Sutskever, Ilya and Cobbe, Karl},
  journal={arXiv preprint arXiv:2305.20050},
  year={2023}
}

@article{zoph2022stmoe,
  title={{ST-MoE}: Designing Stable and Transferable Sparse Expert Models},
  author={Zoph, Barret and Bello, Irwan and Kumar, Sameer and Du, Nan and Huang, Yanping and Dean, Jeffrey and Shazeer, Noam and Fedus, William},
  journal={arXiv preprint arXiv:2202.08906},
  year={2022}
}

@inproceedings{lepikhin2021gshard,
  title={{GShard}: Scaling Giant Models with Conditional Computation and Automatic Sharding},
  author={Lepikhin, Dmitry and Lee, HyoukJoong and Xu, Yuanzhong and Chen, Dehao and Firat, Orhan and Huang, Yanping and Krikun, Maxim and Shazeer, Noam and Chen, Zhifeng},
  booktitle={Proceedings of the International Conference on Learning Representations},
  year={2021}
}

@inproceedings{fu2023moefication,
  title={Go Beyond the Impossible: {MoEfication} of Transformer Models},
  author={Fu, Zhengxiao and Zhang, Qingqing and Liu, Xiao and Liu, Zhiyuan and others},
  journal={arXiv preprint arXiv:2305.07908},
  year={2023}
}

@inproceedings{shazeer2017outrageously,
  title={Outrageously Large Neural Networks: The Sparsely-Gated Mixture-of-Experts Layer},
  author={Shazeer, Noam and Mirhoseini, Azalia and Maziarz, Krzysztof and Davis, Andy and Le, Quoc and Hinton, Geoffrey and Dean, Jeff},
  booktitle={Proceedings of the International Conference on Learning Representations},
  year={2017}
}

@inproceedings{frantar2023sparsegpt,
  title={Sparse{GPT}: Massive Language Models Can Be Accurately Pruned in One-Shot},
  author={Frantar, Elias and Alistarh, Dan},
  booktitle={Proceedings of the 40th International Conference on Machine Learning},
  year={2023}
}

@inproceedings{sun2024simple,
  title={A Simple and Effective Pruning Approach for Large Language Models},
  author={Sun, Mingjie and Liu, Zhuang and Bair, Anna and Kolter, J Zico},
  booktitle={Proceedings of the International Conference on Learning Representations},
  year={2024}
}

\end{document}